# RGBD-based Parameter Extraction for Door Opening Tasks with Human Assists in Nuclear Rescue

Jiajun Li[1], Jianguo Tao[1*], Liang Ding[1], Haibo Gao[1], Zongquan Deng[1], Yu Wu[2]
1: State Key Laboratory of Robotics and System, Harbin Institute of Technology, Harbin, China
2: China Nuclear Power Technology Research Institute, Shenzhen, China
*: jgtao@hit.edu.cn

***Abstract*—The ability to open a door is essential for robots to perform home-serving and rescuing tasks. A substantial problem is to obtain the necessary parameters such as the width of the door and the length of the handle. Many researchers utilize computer vision techniques to extract the parameters automatically which lead to fine but not very stable results because of the complexity of the environment. We propose a method that utilizes an RGBD sensor and a GUI for users to "point" at the target region with a mouse to acquire 3D information. Algorithms that can extract important parameters from the selected points are designed. To avoid large internal force induced by the misalignment of the robot orientation and the normal of the door plane, we design a module that can compute the normal of the plane by pointing at three non-collinear points and then drive the robot to the desired orientation. We carried out experiments on real robot. The result shows that the designed GUI and algorithms can help find the necessary parameters stably and get the robot prepared for further operations.**

## I. Introduction

Robots today are not limited in structural artificial environments like factories. They are required to participate in all kinds of complex daily activities such as home serving and disaster rescuing. These kinds of activities includes manipulate different objects. And an essential problem in robot manipulation is how to get the 3D location of the objects that is going to be operated.

Hai and Charles utilized a green laser pointer to illuminate the task-relevant area so that the robot will have an approximate target position to navigate to [1]. This method is quite convenient in situations where humans are on the spot.

Anna and Oussama tried to get the location of the objects by touching [2]. They regarded the problem as a Bayesian posterior estimation problem and proposed an efficient method to estimate in real time. Their method can be applied without human's assist.

RGBD sensors such as Kinect and Asus Xtion are widely used by researchers in robotics and computer vision due to its low price in recent years. They can receive not only the color images as regular cameras can do, but also the 3D information which used to be hard to acquire. With 3D information of images, we can gain more insights into to the spatial and geometry features.

Thomas and Jurgen designed a model with two basic functional elements: one for detecting specular handles and the other for non-specular handles [3]. They did so because the RGBD sensors use infrared light to get the depth information which may cause problems with specular surfaces. Their approach was tested on PR2 robot and was successful in 51.9% of all 104 trails.

Wim and Melonee carried out the door and handle detection with both RGBD sensors and laser scanners [4]. During the handle detecting process, an object classifier which is implemented by OpenCV's HaarClassifier Cascade is utilized, then the 3D range information from the stereo camera is used to filter out the false positives.

Ellen and Ashutosh [5] proposed a method that tried to locate a door handle with color images and then make use of the 3D range information to extract the "3D key locations" of the handle such as the axis of door handles and the end point of the handles. They used the sliding window object detection method to locate the handles and tuned the empirical constants which indicate the "3D key locations" to get the geometry feature of the handles.

Zhou and Zeyang described their method to detect doors and handles in detail [6]. First they extract the door plane with a RANSAC method [7], then the points lying out of the plane are recognized as handle points. The rotation center and the grasp point of the door handle are found by fitting the points to a cuboid model and regarding the end points on the edge as the target points.

Dmitri and Galina proposed a method to localize the handle with one CCD camera [8]. They extract 3D point cloud using the optical flow calculated from images. With these 3D points, they found the door plane using RANSAC method, then they

This work was supported by the National Key Basic Research Development Plan Project (973) (2013CB035502), Harbin Talent Programme for Distinguished Young Scholars (No. 2014RFYXJ001), Research Project of State Key Laboratory of Mechanical System and Vibration (MSV201610), Fundamental Research Funds for the Central Universities (Grant No. HIT.BRETIII.201411), and "111" Project (B07018).

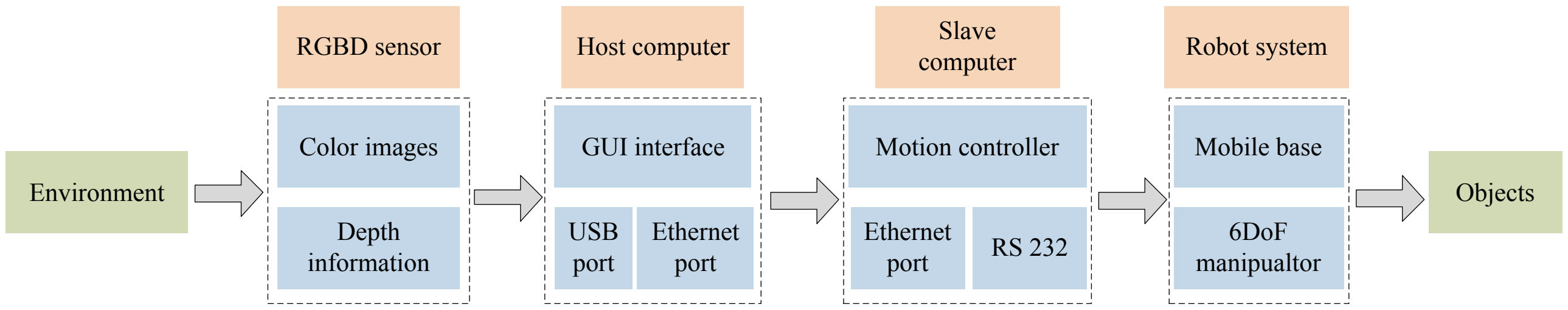

Figure 1. The principle diagram of the vision system

removed the door points by clustering them into one group to get the handle points.

All these described method above use 3D points cloud to detect and localize doors and handles, though these 3D points are obtained in different ways. Their aims are to let the robots find the object which they will operate automatically, without human's supervision. However, in some scenarios like nuclear rescuing, robots have to deal with dynamic and complex environments and high degree of automation like this may cause permanent damage to the robot. In this situation, teleoperation with human assist could be a better option. To make the teleoperation more efficient, operators have to observe the robots and their surroundings carefully. A GUI interface is quite convenient for operators to "pick" the target and direct the robot and RGBD sensors will supply useful 3D information to operators.

In this paper, an RGBD sensor based GUI interface is designed and applied to door opening tasks to help extract the necessary 3D information of the doors and handles. The remainder of this paper will proceed as follow: Section II introduces the RGBD sensor we use and the vision system of the robot. The working process with the RGBD sensor to extract 3D information and open doors is described in Section III. Section IV details each procedure in Section III. In Section V, we propose a method to automatically adjust the orientation of the mobile base to be perpendicular to the door plane. Section VI shows the experimental validation of the proposed methods.

## II. The Stereo Vision System

In this work, we use an ASUS Xtion Pro Live to obtain color and depth images at 30fps, which is enough for real time applications. The 6DoF manipulator is mounted directly on the mobile base as shown in Fig. 2, and the RGBD camera is placed in front of the mobile base so that it can get a wide perspective of the target rather than being blocked by the manipulator. The base frame and the sensor frame are both fixed, so there is a constant transfer matrix between the two which we will discuss in section III.

The vision system principle diagram is shown in Fig. 1. The RGBD sensor is directly connected to the host computer through a USB port so that the time-consuming image processing will not cause delay in slave computer which controls the motion of the robot. The host and the slave computers are connected through Ethernet port. The slave computer controls the 6DoF manipulator via RS 232 and the mobile base via a GoogolTech GTS4 motion control card.

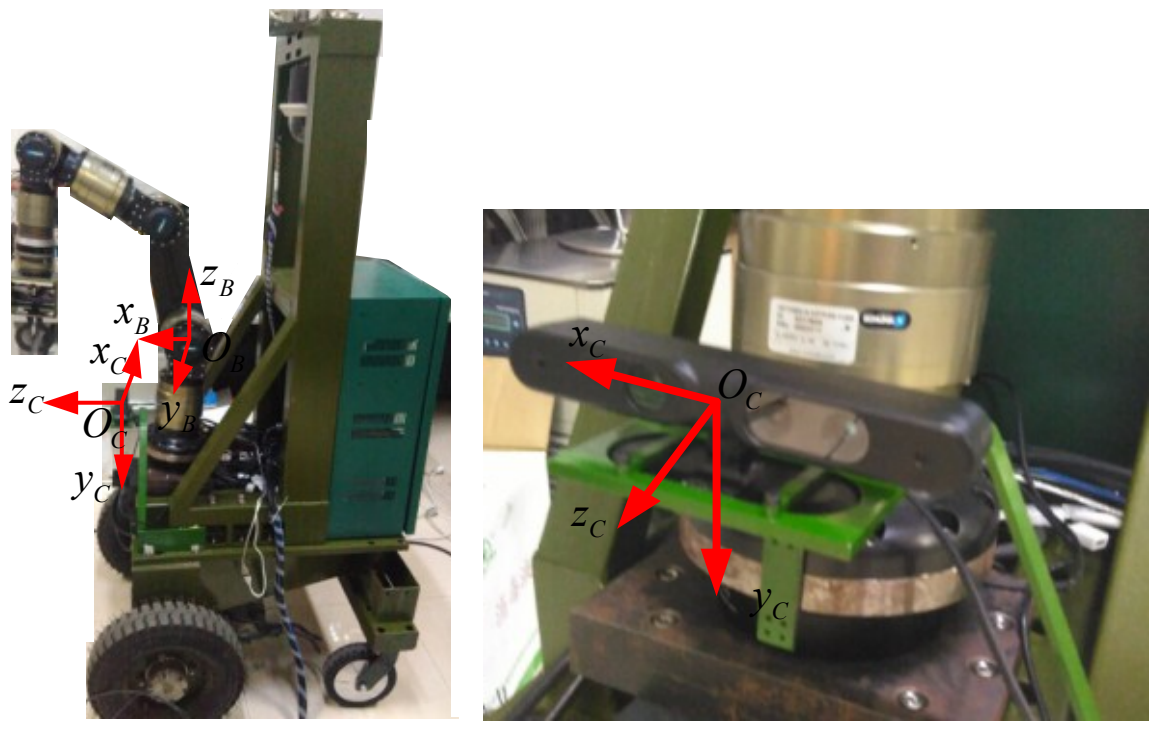

Figure 2. frames of the manipulator and the RGBD camera

## III. The Control Structure Design

In this section, we will briefly introduce the general framework of the designed GUI interface and the standard operational procedure. More details about the GUI are discussed in section IV.

### *A. The Graphical User Interface*

The GUI is programmed in C++ with MFC (VS 2010) on WINDOWS 7 system. We communicate with the ASUS Xtion with OpenNi2 model and then use OpenCV to finish image processing. We will extract the 3D information we need with the help of GUI and pass these parameters to the slave computer via TCP socket. The GUI shown in Fig. 3 consists of four parts: the image displaying region which is also the area where we perform "point to target" actions with the mouse, the parameter extracting region, the real time data displaying region and the command sending region.

The image display region is used to display color images at 30fps. The original data flow including color images and depth information is captured by the host computer. Then we align the depth information with respect to the color images to eliminate the intrinsic offset between the RGB sensor and the Depth sensor. After that we represent the 3D information with the camera coordinate. The functions used in these two steps are provided by OpenNI2.

We represent 3D information in the base frame with (1), where ${}^{B}p$ and ${}^{C}p$ are the coordinates in base frame and in camera frame respectively, ${}^{B}_{C}T$ is the homogeneous transfer matrix between the two frames.

$$ {}^{B}p = {}^{B}_{C}T \cdot {}^{C}p \qquad (1) $$

In this paper, ${}^{B}_{C}T$ is a constant matrix shown in (2):

$$ {}^{B}_{C}T = \begin{bmatrix} 0 & 0 & 1 & 15.0 \\ -1 & 0 & 0 & 1.5 \\ 0 & -1 & 0 & -22.0 \\ 0 & 0 & 0 & 1 \end{bmatrix} \quad (2) $$

Using this transfer matrix we can easily project a 3D point from the camera frame to the base frame. So we designed the real time data displaying region to show the coordinates in the camera and the base frame. Every time the cursor is moved in the image displaying region, the program automatically grub the 3D information and compute the coordinates for the pointed object. This region is useful when we want to get an intuitive feels about how close we are now to the specific objects or obstacles.

The parameter extracting region is used to extract the parameters we need for path planning. After we get the necessary parameters, the motion commands are sent to the slave computer through the buttons in command sending region.

### B. *The standard operation procedure*

The standard operation procedure is shown in Fig. 4. First we have to drive the robot to about 1.0 meter in front of the door so that we can have the left and right edges of the door being detected without much error. The distance between the robot and the door can be controlled by continuously move the cursor on the door plane, so that the distance can be viewed real time displaying region. Then we navigate the robot to get closer to the door and measure the door handle length. The distance is about 0.75 meters. Then the normal of the door plane is extracted and the robot can adjust its orientation to be perpendicular to the normal of the door plane. Finally, the contact point is obtained after the robot has located in a proper place.

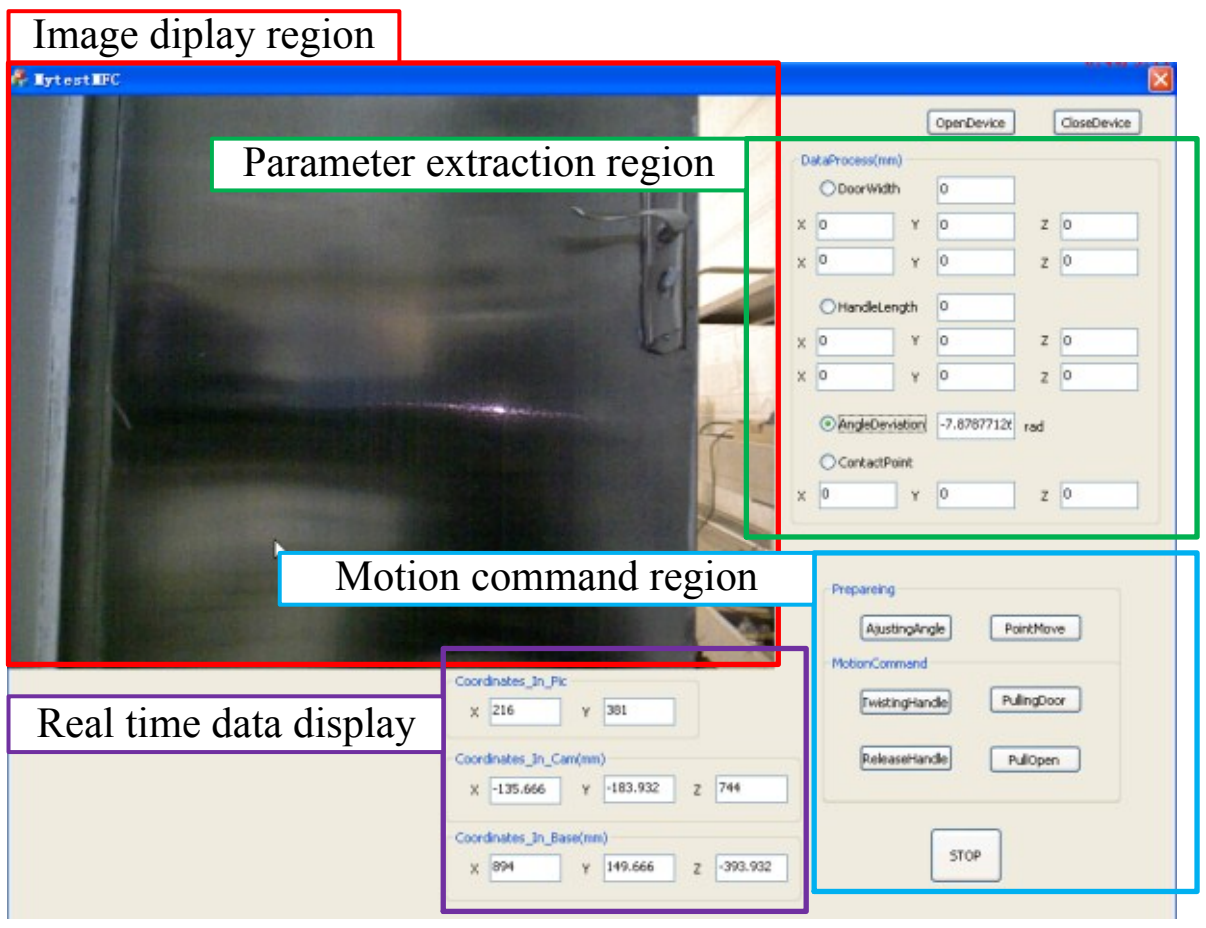

Figure 3. Four functional regions in GUI

With the parameters detected, we send the command to the slave computer and the robot can plan the following motion such as twisting door handles and pulling open doors. The motion planning process is beyond the scope of this paper, and is discussed in [9].

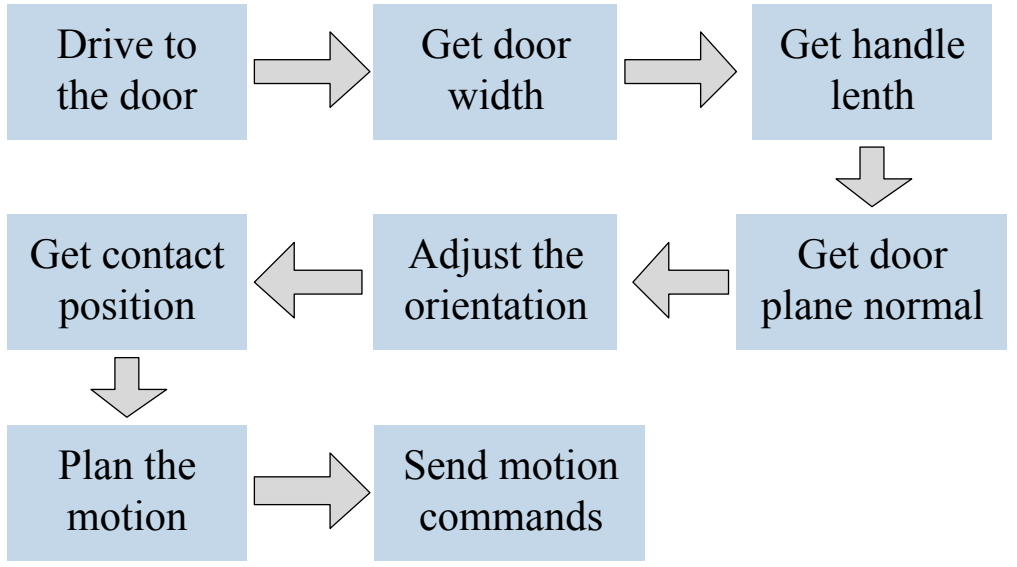

Figure 4. The standard operation procedure

## IV. Parameters Extraction

In the next section, we will focus on the details of how to get different parameters from pointing.

In [9], we discuss the path generation process for door handle twisting and door pulling and package a program written in C++ which takes in three parameters: door width, door handle length, and 3D coordinates of contact point located on the handle and generate path points with time stamps for the manipulator and the mobile base. In this section, we will describe in details how to get the three parameters. Note that in this paper, we only discuss cubic like door handles, not cylindrical handles.

### A. *The width of the door*

When the robot is about 2 meters in front of the door, we start to find the width of the door by pointing at the target area as shown in Fig. 5(a). $P_1$ is the first point we choose which also indicates the position of the door axis of rotation. $P_2$ is the second point which indicates the edge on the moving side. Notice that there is a small offset between $P_2$ and the edge of the door which means the robot operate the handle, not the edge when pulling open doors.

---

Algorithm 1: Dist_Ori_Compute

Input: Two points $P_1(x_1, y_1, z_1)$ and $P_2(x_2, y_2, z_2)$

Output: Distance between the two points $d_{DW}$

Direction of rotation $O_{DR}$

---

1: $d_{DW} \leftarrow \sqrt{(x_1 - x_2)^2 + (y_1 - y_2)^2 + (z_1 - z_2)^2}$

2: If $y_1 - y_2 > 0$ then $O_{DR} \leftarrow$ CCW

3: Else then $O_{DR} \leftarrow$ CW

4: End if

5: Return $d_{DW}$, $O_{DR}$

---

We then compute the width of the door with the two points as shown in Algorithm 1. This algorithm takes in two points and returns the distance between the two points, and the direction of rotation of the door. The direction of rotation is illustrated in Fig. 5(b).

### B. *The length of the handle*

After the width of the door is extracted, the robot is driven to the door. The ideal distance between the robot and the door plane is about 0.75 meters. This data takes the working space of the manipulator and the minimum detecting range of the RGBD sensor into consideration and can be monitored during the driving process.

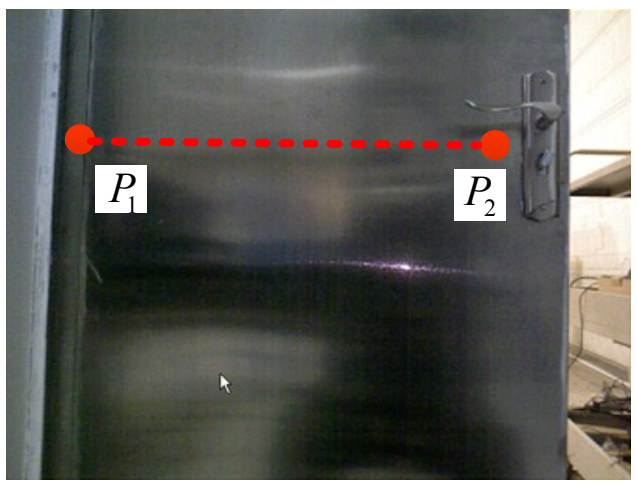

(a) measuring the width of the door

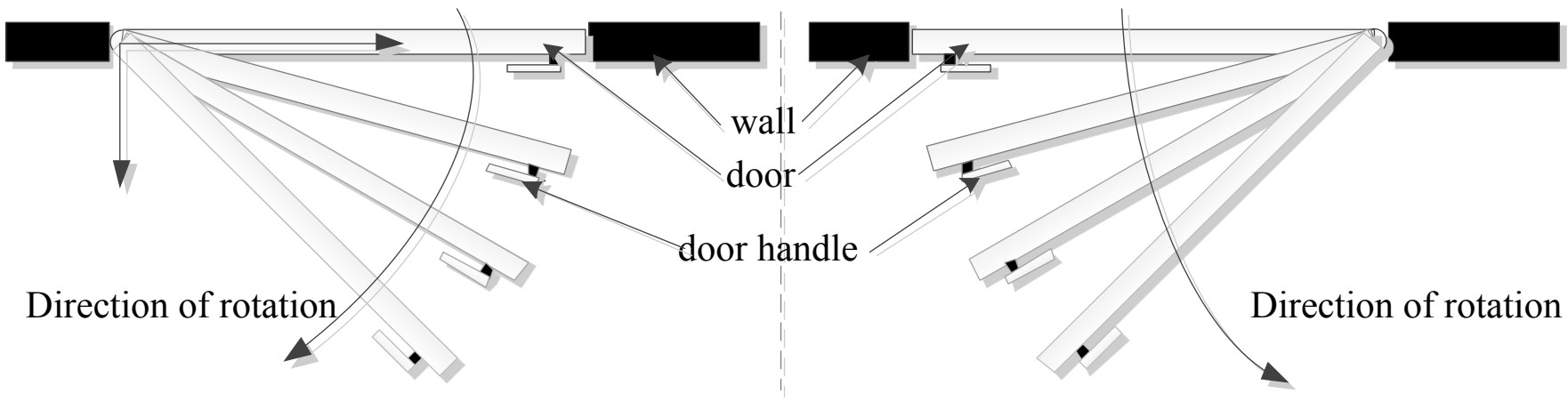

(b) different directions of door opening

Figure 5. The width of a door and the different directions of rotation

The parameter extraction process is basically the same. We pick the first point that indicates the rotation axis of the handle, then the second point on the other edge of the handle. The distance between the two points and the direction of rotation are extracted by Algorithm 1.

### C. *The contact point*

After we get the width of the door and the length of the handle, we start to adjust the robot orientation to be perpendicular to the door plane which we will discuss in section V. The last thing needed to do is to tell the robot where to start. And the contact point on the edge of the handle is the place where the robot starts to twist.

One thing we cannot ignore is that when the object has specular or transparent surface, the RGBD sensor based on infrared ray will become unstable, and holes on the surface with zero feedback will appear. This phenomenon to some extent diminishes the value of the sensor. Luckily, with human aids, this drawback can be avoided by selecting on an area that can be measured and is parallel with the target region.

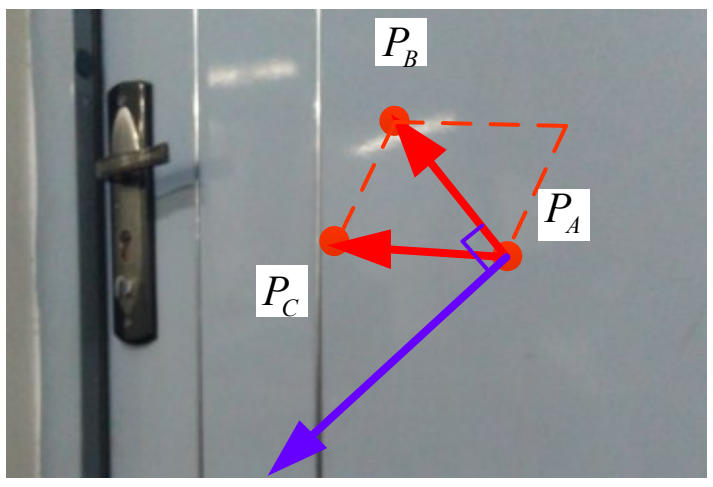

Figure 6. Computing the normal of a door

## V. CONTROL OF THE MOBILE BASE WITH VISUAL AIDS

In this section, we will discuss (a) how to use the GUI to get the normal of the door plane, and (b) how to use the normal to adjust the orientation of the robot. This step should be done before we obtain the contact point.

### A. *The normal of the door plane*

To compute the normal of the door plane, first we have to find the analytical equation of the plane. We randomly select three non-collinear points on the plane with GUI as shown in Fig. 6. Then formulate two vectors from the three points as shown in (3).

$$\begin{aligned} \overrightarrow{P_A P_B} &= [(x_B - x_A),(y_B - y_A),(z_B - z_A)]^T \\ \overrightarrow{P_A P_C} &= [(x_C - x_A),(y_C - y_A),(z_C - z_A)]^T \end{aligned} \quad (3)$$

The two vectors intersecting on point $P_A$ lie in the door plane. Using cross-product, we will find the vector that representing the direction of the normal of the plane as shown in (4).

$$\overrightarrow{P_A P_B} \times \overrightarrow{P_A P_C} = \begin{bmatrix} (y_B - y_A)(z_C - z_A) - (z_B - z_A)(y_C - y_A) \\ (x_C - x_A)(z_B - z_A) - (x_B - x_A)(z_C - z_A) \\ (x_B - x_A)(y_C - y_A) - (x_C - x_A)(y_B - y_A) \end{bmatrix} \quad (4)$$

Let $\vec{n_d}$ denote $\overrightarrow{P_A P_B} \times \overrightarrow{P_A P_C}$. The angle between $\vec{n_d}$ and the y axis of the base frame can be computed using (5):

$$\theta_{diff} = \arccos\left(\frac{(x_C - x_A)(z_B - z_A) - (x_B - x_A)(z_C - z_A)}{\left|\vec{n_d}\right|}\right) \quad (5)$$

### B. *Mobile base controlling*

After we get the deviation angle, we start to drive the mobile base to eliminate the deviation. The mobile base has two driving wheels in the front which can be independently controlled, and two driven wheels at the back. Thus the locomotion system can be view as a differential-driving model as shown in Fig. 7.

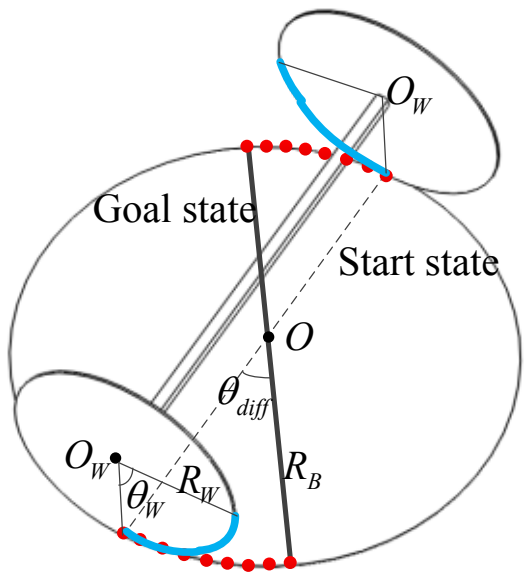

Figure 7. The differential-drive model of the mobile bas

As shown in Fig. 7, the grey dot straight line represents the start state, and the black solid line represents the goal state. The angle between these two states is the $\theta_{diff}$ we have just computed in (5). Suppose there is no slide between the wheels

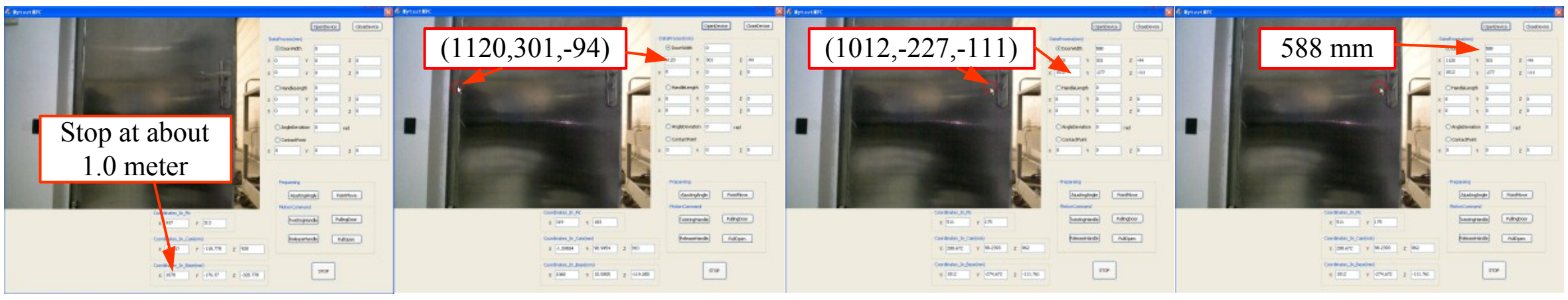

(a) The process of extracting the width of the door

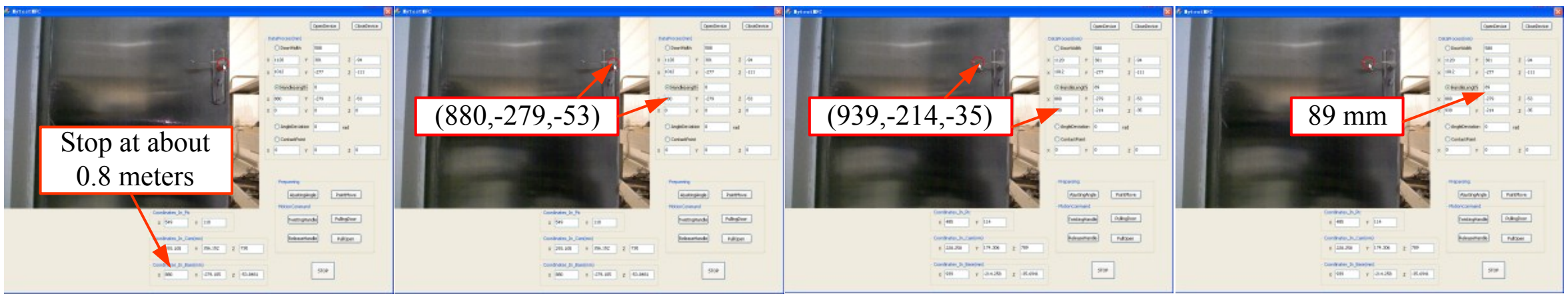

(b) The process of extracting the length of the handle

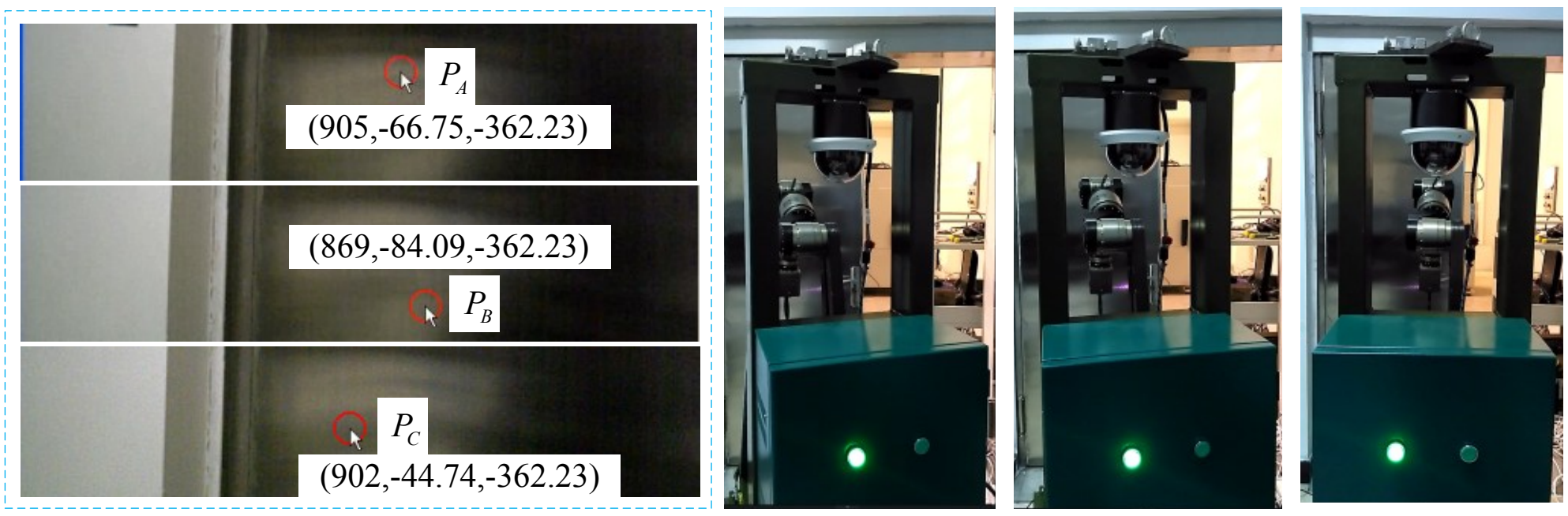

(c) The process of adjusting the orientation of the robot

Figure 8. Snapshots during the experiments

and the ground, the two wheels can rotate a certain angle $\theta_W$ in opposite directions to eliminate the error. To compute $\theta_W$, we need to use the geometry parameters in Fig. 7.

The radius of the wheels and the radius of the rotation with respect to axis $O$ are denoted by $R_W$ and $R_B$. The red dot curves represent the path that the robot has to rotate through, and the blue solid curves are the corresponding arc length. They have the same length, so we have (6):

$$\theta_W R_W = \theta_{diff} R_B \tag{6}$$

Solving equation (6), we will get

$$\theta_W = \frac{\theta_{diff} R_B}{R_W} \tag{7}$$

$\theta_W$ can be sent to the mobile base controller to control the motion of the wheels.

## VI. Experiments

To validate our approach, we carried out experiments on real robot to extract necessary parameters of the door in our lab. We first navigate the robot to about 1.0 meter in front of the door shown in Fig. 8 (a), and get the width of the door. Then move the robot forward to get closer to the door and obtain the length of the handle as shown in Fig. 8 (b). Adjusting the orientation of the robot to be perpendicular to the door plane is shown in Fig. 8 (c). The last step is pointing to the contact point and sending motion command to the robot.

## VII. Conclusion

In this paper, we present our robot vision system based on an RGBD sensor for extracting necessary parameters with human assists, such as the width of the door and the length of the door handle during the pre-operation process of door opening. The extracted parameters can be sent to the slave computer for motion planning. A GUI is design to help visualize the extracting and sending process. Users can directly point at the desired target region, and the GUI program will find the corresponding 3D coordinates. The width of the door and the length of the handle can be computed automatically after two points representing the start and end points are selected. We also design an orientation adjust module for the mobile base in case that the direction the robot is facing is not perpendicular to the normal of the door plane. The proposed GUI and module are tested on a real robot, and the results show that the parameters can be extracted and the robot's orientation can be easily adjusted.